\pdfoutput=1

\documentclass[11pt]{article}

\usepackage[preprint]{acl}

\usepackage{times}
\usepackage{latexsym}

\usepackage{scalefnt}
\usepackage{amsfonts}
\usepackage{amsmath,bm}
\usepackage{mathrsfs}
\usepackage{multirow}
\usepackage{verbatim}
\usepackage{amssymb}
\usepackage{booktabs}
\usepackage{subcaption}
\usepackage{graphicx}
\usepackage{colortbl}
\usepackage{xspace}
\usepackage{CJKutf8}

\usepackage[T1]{fontenc}

\usepackage[utf8]{inputenc}

\usepackage{microtype}

\usepackage{inconsolata}

\newcommand{\model}[0]{\textsc{CritiqueLLM}\xspace}
\newcommand{\blanksymbolfootnote}[1]{%
  \renewcommand{\thefootnote}{}
  \footnote{#1}%
  \setcounter{footnote}{0} 
  \renewcommand{\thefootnote}{\arabic{footnote}}
}

\title{\textsc{CritiqueLLM}: Towards an Informative Critique Generation Model for Evaluation of Large Language Model Generation}

\author{Pei Ke$^{1,*}$, Bosi Wen$^{1,2,*,\dagger}$, Zhuoer Feng$^{1,2,*,\dagger}$, Xiao Liu$^{3,2,*}$, \\
\textbf{Xuanyu Lei$^{3,2,\dagger}$, Jiale Cheng$^{1,2,\dagger}$, Shengyuan Wang$^{3,2,\dagger}$, Aohan Zeng$^{3,2,\dagger}$,}\\
\textbf{Yuxiao Dong$^{3}$, Hongning Wang$^{1}$, Jie Tang$^{3}$, Minlie Huang$^{1,\ddagger}$} \\
$^1$The Conversational Artificial Intelligence (CoAI) Group, Tsinghua University \\
$^2$Zhipu AI\\
$^3$The Knowledge Engineering Group (KEG), Tsinghua University\\
\tt\small kepei1106@outlook.com, \ \{wbs23,fze22,liuxiao21\}@mails.tsinghua.edu.cn, aihuang@tsinghua.edu.cn \\}

\begin{document}
\maketitle

\blanksymbolfootnote{$^*$Equal contribution}
\blanksymbolfootnote{$^\dagger$Work done when these authors interned at Zhipu AI.}
\blanksymbolfootnote{$^\ddagger$Corresponding author}

\begin{abstract}


Since the natural language processing (NLP) community started to make large language models (LLMs) act as a critic to evaluate the quality of generated texts, most of the existing works train a critique generation model on the evaluation data labeled by GPT-4's direct prompting. We observe that these models lack the ability to generate informative critiques in both pointwise grading and pairwise comparison especially without references. As a result, their generated critiques cannot provide fine-grained distinguishability on generated texts, causing unsatisfactory evaluation performance. In this paper, we propose a simple yet effective method called \textit{Eval-Instruct}, which can first acquire pointwise grading critiques with pseudo references and then revise these critiques via multi-path prompting to obtain informative evaluation data in different tasks and settings, including pointwise grading and pairwise comparison with / without references. After fine-tuning on these data, the resulting model \model is empirically shown to outperform ChatGPT and all the open-source baselines and even achieve comparable evaluation performance to GPT-4 in system-level correlations of pointwise grading. We also demonstrate that our generated critiques can act as scalable feedback to further improve the generation quality of strong LLMs like ChatGPT\footnote{The codes are available at \url{https://github.com/thu-coai/CritiqueLLM}.}.

\end{abstract}

\section{Introduction}

Recently, large language models (LLMs) \cite{openai2022chatgpt,openai2023gpt4,touvron2023llama} have been improved rapidly and approached human-level performance on various natural language processing (NLP) tasks, such as question answering, text summarization, dialogue generation, and code generation \cite{laskar2023systematic}. How to automatically measure the performance of LLMs has now become an essential research problem and attracted extensive attention \cite{chang2023llmevalsurvey,zhang2023safetybench,liu2023agentbench}. Strong evaluation methods are expected to provide high-quality critiques (including not only rating scores but also explanations) that act as scalable feedback and guide LLMs to improve persistently \cite{cui2023ultrafeedback}.

Traditional evaluation metrics, usually based on n-gram overlap between generated texts and reference texts (such as BLEU \cite{papi2002bleu} and ROUGE \cite{lin2004rouge}), have limited effectiveness. Recent works mostly resort to model-based evaluation metrics, especially LLM-based ones \cite{wang2023chatgptnlgeval,liu2023geval,zheng2023judging}. 
Since most of the best-performing LLMs such as ChatGPT \cite{openai2022chatgpt} and GPT-4 \cite{openai2023gpt4} can only be accessed via OpenAI APIs, researchers start to automatically collect evaluation data by directly prompting GPT-4 and train their own evaluation models, aiming to avoid potential risks of commerical APIs, such as high cost, unstable usage, and data leakage \cite{zheng2023judging,wang2023pandalm,li2023autoj}.



However, we argue that these evaluation models are still struggling to generate informative critiques in different evaluation tasks including pointwise grading and pairwise comparison. Especially in the challenging reference-free setting, these models tend to generate general critiques without fine-grained distinguishability on generated texts, causing unsatisfactory evaluation performance \cite{zheng2023judging}.

In this work, we propose a simple yet effective method called \textit{Eval-Instruct}, which can automatically construct informative instruction-tuning data for different evaluation tasks and settings, including pointwise grading and pairwise comparison with / without references. Our main idea is to fully utilize referenced pointwise grading critiques, which are shown to possess rich information
with the assistance of references and elaborate prompt design \cite{zheng2023judging,lei2023alignbench}, to construct evaluation data for other tasks and settings. Specifically, after acquiring pointwise grading critiques with pseudo references via GPT-4, we devise a multi-path prompting method including two strategies: 
1) \textbf{Pointwise-to-Pairwise} Prompting aims to inject pointwise grading critiques into pairwise critiques and enrich them with more information about the respective quality of text pairs.
2) \textbf{Referenced-to-Reference-Free} Prompting is targeted at removing direct comparison with references in referenced critiques, while keeping other details to improve the specificity of reference-free critiques.
The evaluation data in different tasks and settings can be acquired via different paths consisting of these two strategies. And we also design a cross validation mechanism to improve the data quality of reference-free pairwise comparison because both of the two paths reach this task.
After fine-tuning on the data of all the tasks and settings, the resulting model \model is empirically shown to outperform all the open-source baselines and even achieve comparable performance with GPT-4 in system-level correlations of pointwise grading.
We also show the potential of \model to act as effective feedback to enhance the performance of LLMs like ChatGPT.


Our main contributions are as follows:
\begin{itemize}
    \item We propose an evaluation data construction method called Eval-Instruct to automatically acquire informative evaluation data in both pointwise grading and pairwise comparison with / without references. 
    \item We conduct extensive experiments on \model, which is fine-tuned on the data constructed by Eval-Instruct. Experimental results on three instruction following benchmark datasets show that our model can outperform all the open-source baselines and even perform comparably with GPT-4 in system-level correlations of pointwise grading.
    \item We reveal the potential of \model to guide LLMs to improve persistently by showing the positive impact of our generated critiques as scalable feedback on the generation quality of LLMs.
\end{itemize}

\section{Related Work}

Evaluation is a long-standing task in NLP, which becomes more challenging with the rapid development of LLMs \cite{celi2020evalsurvey,chang2023llmevalsurvey}. Currently, there are mainly two lines of work on LLM evaluation, including NLU-style and NLG-style evaluations. NLU-style evaluation methods utilize natural language understanding (NLU) tasks such as multi-choice QA to measure the performance of LLMs via simple objective metrics (such as accuracy and F1 score) \cite{dan2021mmlu,zhong2023agieval,huang2023ceval}, 
which may deviate from the common usage of LLMs and may not exactly reflect the ability of LLMs in generating responses for user queries. 

NLG-style evaluation methods extend metrics for natural language generation (NLG) tasks and expect to apply them to the measurement of LLM's performance, which are the main focus of this paper.
Compared with early metrics that depend on the n-gram overlap between generated texts and reference texts \cite{papi2002bleu,banerjee2005meteor,lin2004rouge}, recently proposed metrics based on state-of-the-art LLMs like GPT-4 \cite{openai2023gpt4}
are shown to be strong evaluators due to the encouraging effectiveness of LLMs and the simplicity of formulating evaluation tasks as instruction-following tasks \cite{wang2023chatgptnlgeval,chen2023exploring,liu2023geval,zheng2023judging,ke2023decompeval,fu2023gptscore}. 
Since most of the state-of-the-art LLMs can only be accessed via APIs, researchers start to automatically collect evaluation data by directly prompting GPT-4 and train their own evaluation models to provide stable and effective evaluations at a lower cost \cite{wang2023pandalm,li2023autoj,kim2023prometheus}.

The concurrent works similar to ours are the LLMs specially trained for evaluation tasks like PandaLM \cite{wang2023pandalm}, JudgeLM \cite{zhu2023judgelm}, and AUTO-J \cite{li2023autoj}. For comparison, our work is the first attempt to deal with the challenge of uninformative critique generation which commonly appears in recent LLM-based evaluation models especially without references. Instead of prompting GPT-4 directly, our proposed Eval-Instruct can fully utilize the connection among different evaluation tasks and settings to construct informative evaluation data, which are empirically shown to improve the quality of generated critiques.





\section{Method}

\begin{figure*}[!h]
\scriptsize
    \centering
    \includegraphics[width=1.0\textwidth]{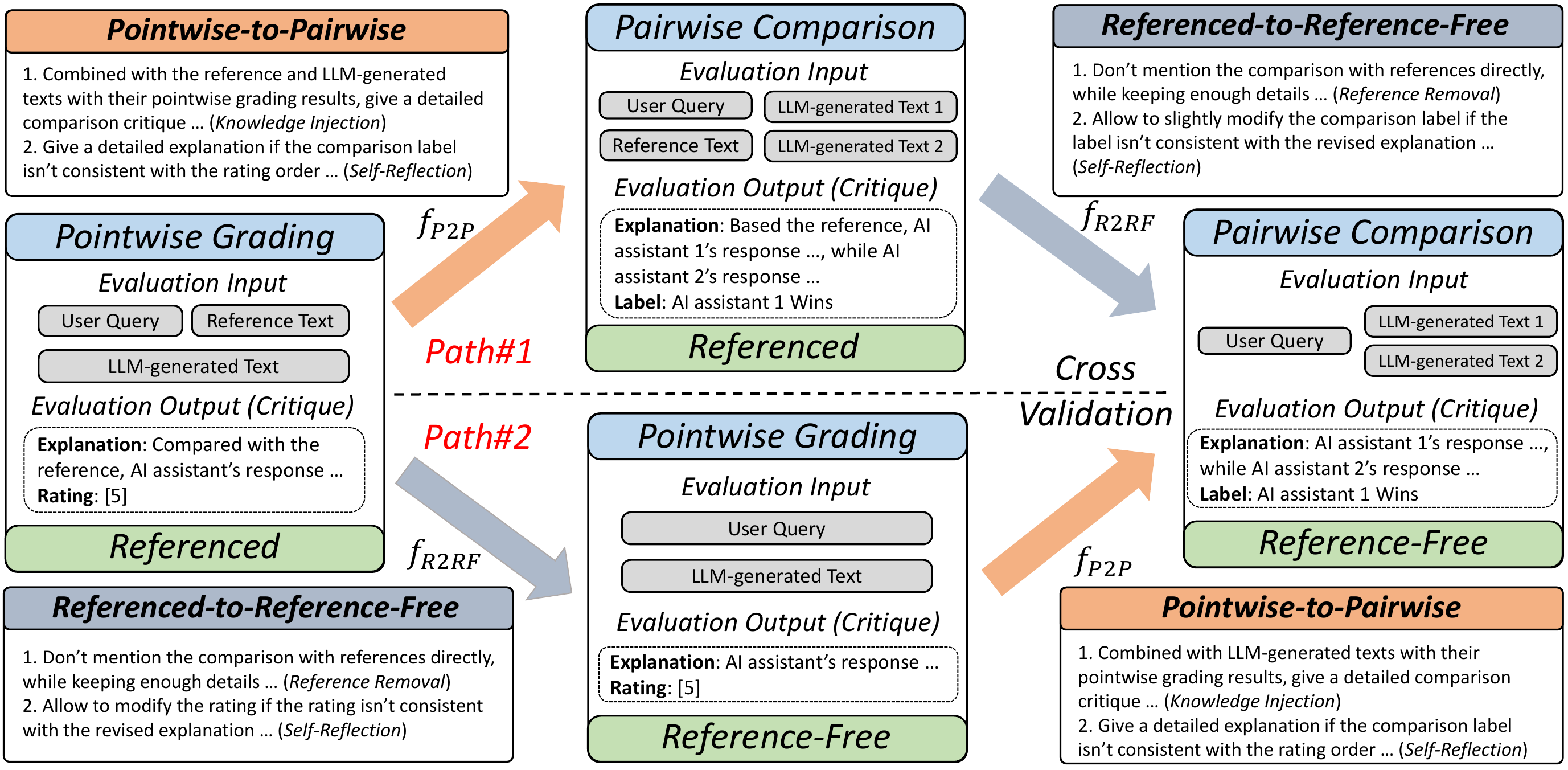}
    \caption{Overview of Eval-Instruct. Starting from referenced pointwise grading data, our proposed multi-path prompting method can apply pointwise-to-pairwise and referenced-to-reference-free prompting strategies to acquire evaluation data in other tasks and settings via two different paths. Cross validation is adopted to filter out the contradictory data from these two paths and further improve the data quality.} 
    \label{fig:overview}
\end{figure*}

\subsection{Task Definition and Method Overview}
\label{sec:taskdef}


This paper mainly involves two typical evaluation tasks:
1) \textbf{Pointwise Grading}: Given a user query $q$, a LLM-generated text $x$, and a reference text $r$ (omitted in the reference-free setting), the goal is to obtain a critique $c$ including a rating score and an explanation to support this score.
2) \textbf{Pairwise Comparison}: Given a user query $q$, two LLM-generated texts $x_1$ and $x_2$, and a reference text $r$ (omitted in the reference-free setting), our purpose is to acquire a critique $c$ including a comparison label (i.e., win / tie / lose) and an explanation to support this label.



Our method consists of the following steps. 
We first construct an informative instruction-tuning dataset for different evaluation tasks and settings, including pointwise grading and pairwise comparison with / without references (\S\ref{sec:traindata}). Specifically, after collecting user queries, LLM-generated texts, and pseudo references (\S\ref{sec:pseref}), 
we can acquire high-quality referenced pointwise grading critiques via elaborately prompting GPT-4. Then, we devise a multi-path prompting method to construct informative evaluation data in other tasks and settings, which covers pointwise-to-pairwise and referenced-to-reference-free prompting strategies (\S\ref{sec:multipath}). Since there are two paths to obtain reference-free pairwise comparison data, we design a cross validation mechanism to filter out the contradictory data and improve the quality (\S\ref{sec:crossvalid}). Finally,
we perform supervised fine-tuning on the automatically constructed evaluation data in a multi-task manner to train a unified critique generation model for different evaluation tasks and settings (\S\ref{sec:sft}).

\subsection{Evaluation-Oriented Instruction Data Construction (\textit{Eval-Instruct})}
\label{sec:traindata}

\subsubsection{Pseudo Reference Collection}
\label{sec:pseref}

To construct instruction-tuning data for evaluation, it is imperative to first obtain the evaluation input, including user queries, LLM-generated texts, and references. We refer to recent works on instruction following \cite{lei2023alignbench,li2023autoj,zhang2023llmeval} and merge their task taxonomy to consider ten instruction following tasks covering diverse NLP applications in real-world scenarios\footnote{Our task taxonomy contains fundamental language ability, advanced Chinese understanding, open-ended question answering, writing ability, logical reasoning, mathematics, task-oriented role play, professional knowledge, code generation, and multi-lingual ability.}.
We utilize self-instruct \cite{wang2023selfinstruct} to augment seed queries of these tasks which are publicly available and conduct strictly filtering to improve the data quality. The details are provided in Appendix \ref{augmentation prompt}.

Then, we collect LLM-generated texts from 10 representative models, which cover different levels of generation qualities,
including GPT-4 \cite{openai2023gpt4}, ChatGPT \cite{openai2022chatgpt}, two versions of ChatGLM \cite{du2022glm,zeng2022glm}, MOSS \cite{sun2023moss}, Minimax\footnote{\url{https://api.minimax.chat/}}, Sparkdesk\footnote{\url{https://xinghuo.xfyun.cn/}}, Chinese-Llama2-7B-Chat\footnote{\url{https://huggingface.co/FlagAlpha/Llama2-Chinese-7b-Chat/}}, Baichuan2-13B-Chat \cite{yang2023baichuan2}, and Ernie Bot\footnote{\url{https://yiyan.baidu.com/}}. We further filter out the generated results by removing a small number of failure cases, such as empty responses. 

Finally, 
we select the best-performing LLM (i.e., GPT-4) and manually check its generated texts for each user query, while revising them if necessary to improve the quality. Thus, these generated texts after manual check and revise can act as 
pseudo references to assist the evaluation data construction.

\subsubsection{Multi-Path Prompting}
\label{sec:multipath}

To acquire high-quality evaluation data in different evaluation tasks and settings, we first construct referenced pointwise grading critiques by prompting GPT-4 with the assistance of pseudo references and well-designed prompts like \citet{lei2023alignbench}, which are empirically shown to be informative \cite{zheng2023judging}. Then, regarding this setting as a beginning, we devise a multi-path prompting method to obtain evaluation data in other tasks and settings. 
As shown in Figure \ref{fig:overview}, there are two main prompting strategies:

\noindent (1) \textbf{Pointwise-to-Pairwise} Prompting ($f_{P2P}$): This prompting strategy injects pointwise grading critiques of generated texts into 
pairwise comparison critiques, enriching them with information about the respective text quality. Meanwhile, it requires self-reflection on the pointwise critiques generated by GPT-4 before obtaining the final pairwise comparison results.

\noindent (2) \textbf{Referenced-to-Reference-Free} Prompting ($f_{R2RF}$): This prompting strategy aims to remove direct comparison with references while keeping informative contents from references. It also requires GPT-4 to self-reflect\footnote{The purpose of self-reflection in the two strategies is to alleviate the inconsistency problem in the output critiques, reducing error propagation during the data construction process.} whether the evaluation results including scores / labels and revised explanations are consistent, and modify the results if necessary.


Equipped with the above prompting strategies, we have two paths to construct evaluation data in different tasks and settings. Assume that $D^{point,r}=\{(q_i,r_i,x_i,c_i^{point,r})\}_{i=1}^N$ indicates the referenced pointwise grading dataset constructed above and $c_i^{point,r}$ represents the critique in the corresponding setting, our purpose is to acquire the datasets $D^{pair,r},D^{point,rf},D^{pair,rf}$ via different paths, where $point/pair$ means pointwise / pairwise evaluation and $r/rf$ indicates referenced / reference-free evaluation, respectively. The two paths are devised as follows.

\noindent \textbf{Path\#1}: $D^{point,r}\xrightarrow{f_{P2P}} D^{pair,r}\xrightarrow{f_{R2RF}} D^{pair,rf}$

As shown in Path\#1 of Figure \ref{fig:overview}, we firstly conduct pointwise-to-pairwise prompting to acquire the referenced pairwise comparison dataset $D^{pair,r}=\{(q_i,r_i,x_{i,1},x_{i,2},c_{i}^{pair,r})\}_{i=1}^M$:
\begin{align}
    c_{i}^{pair,r} & = f_{P2P}(q_i,r_i,x_{i,1},x_{i,2},c_{i,1}^{point,r},c_{i,2}^{point,r}) \notag \\
    i & = 1,2,\cdots,M
\end{align}
%
where $q_i,r_i,x_{i,1},x_{i,2}$ indicate the user query, the reference, and two generated texts of the $i$-th data, respectively. $c_{i,1}^{point,r},c_{i,2}^{point,r},c_{i}^{pair,r}$ are the referenced pointwise and pairwise evaluation results of $x_{i,1},x_{i,2}$, respectively\footnote{We conduct strictly rule-based filtering after each prompting step to remove low-quality data with errors in format and other aspects, which is omitted in this subsection.}. Then, we can apply referenced-to-reference-free prompting to obtain $D^{pair,rf}=\{(q_i,x_{i,1},x_{i,2},c_{i}^{pair,rf})\}$:
%
\begin{align}
    c_{i}^{pair,rf,1} & = f_{R2RF}(q_i,r_i,x_{i,1},x_{i,2},c_{i}^{pair,r}) \notag \\
    i & =1,2,\cdots, M
\end{align}
where $c_{i}^{pair,rf,1}$ means the reference-free pairwise comparison critique of the $i$-th data from Path\#1.

\noindent \textbf{Path\#2}: $D^{point,r}\xrightarrow{f_{R2RF}} D^{point,rf}\stackrel{f_{P2P}}{\rightarrow} D^{pair,rf}$

Similarly, as shown in Path\#2 of Figure \ref{fig:overview}, we can exchange the order of two prompting strategies applied to $D^{point,r}$ accordingly. In this way, we can in turn acquire $D^{point,rf}$ and $D^{pair,rf}$:
\begin{align}
    c_{i}^{point,rf} &  =f_{R2RF}(q_i, r_i,x_i,c_i^{point,r}) \notag \\
    i &  =1,2,\cdots, N \\
c_{i}^{pair,rf,2} & = f_{P2P}(q_i,x_{i,1},x_{i,2},c_{i,1}^{point,rf}, c_{i,2}^{point,rf}) \notag \\
    i & = 1,2,\cdots, M
\end{align}
where $c_{i}^{pair,rf,2}$ denotes the reference-free pairwise comparison critique of the $i$-th data from Path\#2.

\subsubsection{Cross Validation}
\label{sec:crossvalid}

Since both of the two paths finally reach $D^{pair,rf}$, we design a cross validation mechanism to further improve the data quality. Specifically, $D^{pair,rf}$ only contains the data whose comparison labels from two paths are consistent. In this case, the critiques from both of the two paths are added to $D^{pair,rf}$. The other data with contradictory comparison labels are strictly filtered. In our experiment, the proportion of the evaluation data which are filtered out is 7.7\%, demonstrating that most of our constructed data from the two paths have consistent comparison labels, indicating acceptable data quality.

\subsection{Supervised Fine-Tuning}
\label{sec:sft}



We perform supervised fine-tuning on the LLM $P_{\theta}$ using all the constructed training data in a multi-task manner to obtain \model:
\begin{align}
\mathcal{L}  = & -\frac{1}{N}\sum_{i=1}^{N}P_{\theta}(c_i^{point,r}|q_i,r_i,x_i) \notag \\
 & -\frac{1}{N}\sum_{i=1}^{N}P_{\theta}(c_i^{point,rf}|q_i,x_i) \notag \\
 & -\frac{1}{M}\sum_{i=1}^{M}P_{\theta}(c_i^{pair,r}|q_i,r_i,x_{i,1},x_{i,2}) \notag \\
 & -\frac{1}{M^{'}}\sum_{i=1}^{M^{'}}P_{\theta}(c_i^{pair,rf}|q_i,x_{i,1},x_{i,2}) \notag
\end{align}
where $M^{'}$ indicates the data amount of $D^{pair,rf}$ after cross validation. During fine-tuning, we follow \citet{bai2022constitutional} to add simplified prompts to distinguish different parts of inputs. We also follow \citet{li2023autoj} to augment pairwise training data via swapping the order of two generated texts and exchanging the corresponding contents in critiques.

\section{Experiment}

\begin{table} [!t]
\centering
\scriptsize
\setlength{\tabcolsep}{0.7mm}{
\begin{tabular}{l|c|c|c|c|c}
\toprule
Dataset & Task & Setting & \#Models & \#Samples / \#Pairs & Length \\
\midrule
\multirow{2}*{AlignBench} & Pointwise & R / R-F  & 8 & 3,200  &  274 \\ 
& Pairwise  & R / R-F   & 8  & 1,600  & 293 \\ 
\midrule
AUTO-J (Eval-P) & Pairwise  & R-F   & 6 & 1,392  &  372 \\ 
\midrule
LLMEval & Pairwise  & R-F &  11 & 1,530  &  283 \\ 
\bottomrule
\end{tabular}}
\caption{Statistics of the benchmark datasets, including the evaluation task / setting, the number of models / samples / pairs, and the average length of generated texts. R / R-F indicates referenced / reference-free evaluation, respectively.}
\label{tab:datastat}
\end{table}

\subsection{Dataset}
\label{sec:dataset}

We adopt three benchmark datasets on open-ended instruction following, 
which involve various NLP tasks in LLM's real-world scenarios\footnote{We have conducted string matching to show that there is no overlap between the queries in the training and test sets.}. The datasets also cover all the evaluation tasks and settings in this paper. The statistics are shown in Table \ref{tab:datastat}.

\noindent \textbf{AlignBench} \cite{lei2023alignbench}: This benchmark includes 8 categories of instruction following tasks and 8 LLMs for generation.
It provides an evaluation dataset with human-annotated scores on the quality of generated texts. 
In addition to using human-annotated scores for measuring pointwise grading performance, we also follow the original paper to sample text pairs of the same query for pairwise comparison\footnote{The authors in the original paper of AlignBench \cite{lei2023alignbench} collect all the pairs of generated texts for each query ($\sim$10,000 pairwise comparison data), causing high demand of computational resources and API costs for LLM-based evaluation methods. Thus, we randomly sample a subset ($\sim$1,000 pairwise comparison data) to test our method and all the baselines for a fair comparison.}, whose label is automatically determined by their pointwise scores.

\noindent \textbf{AUTO-J (Eval-P)} \cite{li2023autoj}: This benchmark provides 1,392 pairwise comparison data, each of which contains a user query, two LLM-generated texts, and a human-annotated preference label. These data involve 58 real-world scenarios and 6 model families for generation.


\begin{table*} [!h]
\centering
\small
\setlength{\tabcolsep}{1.6mm}{
\begin{tabular}{l|c|c|c|c|c|c|c|c|c|c|c|c}
\toprule
Level & \multicolumn{6}{c|}{Text-Level} & \multicolumn{6}{c}{System-Level} \\
\midrule
Setting  & \multicolumn{3}{c|}{Referenced}  & \multicolumn{3}{c|}{Reference-Free} & \multicolumn{3}{c|}{Referenced}  & \multicolumn{3}{c}{Reference-Free}  \\
\midrule
Metric  & $r$ & $\rho$ & $\tau$ & $r$ & $\rho$ & $\tau$ & $r$ & $\rho$ & $\tau$ & $r$ & $\rho$ & $\tau$ \\
\midrule
\multicolumn{13}{c}{\textit{Closed-Source Evaluation Models}} \\
\midrule
ChatGPT &  0.443 & 0.421  & 0.379  & 0.292  &  0.287 &  0.266  &   0.955  & 0.976   & 0.929  &  0.778  & 0.833  &  0.643 \\
GPT-4 &  \underline{0.629} & \underline{0.583}  &  \underline{0.532} &  \underline{0.523} & \underline{0.494}  & \underline{0.447}   &  \underline{0.995}   &  \underline{1.000}  & \underline{1.000}  &  \underline{0.997}  & \underline{0.976}  & \underline{0.929}  \\
\midrule
\multicolumn{13}{c}{\textit{Open-Source Evaluation Models}} \\
\midrule
ChatGLM3-6B &  0.223 & 0.222  & 0.207  & 0.159  & 0.150  &  0.140  &   0.790  & 0.833   & 0.643  &  0.544  & 0.548  & 0.429 \\
Baichuan2-13B-Chat &  0.199 & 0.200  & 0.187  & 0.125  &  0.117 & 0.110   &  0.854   &  0.929  & 0.786  &  0.663  & 0.527  & 0.400  \\
Qwen-14B-Chat &  0.373 & 0.379  & 0.358  &  0.255 & 0.254  & 0.239   &  0.901   & 0.929   & 0.786  & 0.772   & 0.833  & 0.643  \\
Mixtral-8x7B & 0.474  & 0.471  & 0.426  &  0.302 & 0.306  &  0.282  &  0.972   & 0.976   & 0.929  &  0.863  & 0.929  & 0.786  \\
Llama-2-70B-Chat & 0.152  & 0.162  & 0.109  &  0.123 & 0.122  &  0.113  &  0.663   & 0.667   & 0.500  &  0.547  & 0.429  & 0.286  \\
JudgeLM-13B & 0.450  & 0.430  & 0.391  &  0.170 & 0.162  & 0.155   &  0.984   &  0.976  &  0.929 &  0.717  & 0.905  & 0.786  \\
AUTO-J-Bilingual-6B & -  & -  &  - & 0.044  & 0.045  & 0.041   &  -   & -   & -  &  0.558  & 0.571  & 0.500  \\
\midrule
\model (Ours) & \textbf{0.555}  &  \textbf{0.523} & \textbf{0.477}  & \textbf{0.366}  &  \textbf{0.352} & \textbf{0.319}   &  \underline{\textbf{0.995}} & \underline{\textbf{1.000}}   & \underline{\textbf{1.000}}  & \textbf{0.954}   & \underline{\textbf{0.976}}  & \underline{\textbf{0.929}}   \\
\bottomrule
\end{tabular}}
\caption{Text-level and system-level Pearson ($r$), Spearman ($\rho$), and Kendall ($\tau$) correlations in referenced and reference-free settings of pointwise grading on AlignBench. The highest correlation among the methods based on local models is \textbf{bold}, while the highest correlation overall is \underline{underlined}. - means that AUTO-J-Bilingual-6B cannot support referenced pointwise grading.}
\label{tab:mainsum}
\end{table*}

\noindent \textbf{LLMEval} \cite{zhang2023llmeval}: This benchmark designs 17 types of user queries covering representative NLP tasks in real-world scenarios, and provides $\sim$ 100,000 pairwise comparison data with human-annotated labels. Due to the limitation of computational resources and API costs for LLM-based evaluation methods, we randomly sample a subset ($\sim$1,000) to measure the performance of our method and all the baselines for a fair comparison.

As for the relationship between our training dataset in \S\ref{sec:traindata} and these benchmark datasets, our training dataset has similar task categories with AlignBench because our task taxonomy is built mainly based on AlignBench \cite{lei2023alignbench} with other tasks in recent works \cite{li2023autoj,zhang2023llmeval} as supplementary, as described in \S\ref{sec:pseref}. Also, our training dataset includes the training data of AUTO-J (Eval-P) \cite{li2023autoj} while excluding its test set. Compared with AlignBench and AUTO-J (Eval-P), LLMEval \cite{zhang2023llmeval} does not have a similar task or data distribution with our training dataset, which can act as a benchmark to test the generalization ability.



\subsection{Baselines}


We choose state-of-the-art general LLMs and evaluation-specific LLMs as our baselines.

\noindent \textbf{General LLMs}: We adopt ChatGPT (\texttt{gpt-3.5-turbo-1106}) \cite{openai2022chatgpt}, GPT-4 (\texttt{gpt-4-1106-preview}) \cite{openai2023gpt4}, ChatGLM3-6B \cite{du2022glm,zeng2022glm}, Baichuan2-13B-Chat \cite{yang2023baichuan2}, Qwen-14B-Chat \cite{bai2023qwen}, Llama-2-70B-Chat \cite{touvron2023llama2}, and Mixtral-8x7B \cite{jiang2024mixtral} as our general baselines.
These general LLMs can perform as an evaluator for pointwise grading and pairwise comparison via elaborate prompts without further training. 
We directly prompt these LLM to obtain evaluation results in single-turn interaction.

\noindent \textbf{Evaluation-Specific LLMs}: We select AUTO-J-Bilingual-6B \cite{li2023autoj} and JudgeLM-13B \cite{zhu2023judgelm} as our task-specific baselines. These two baselines are designed for specific evaluation tasks and settings.


\begin{table*} [!h]
\centering
\small
\begin{tabular}{l|c|c|c|c|c|c|c|c}
\toprule
Dataset & \multicolumn{4}{c|}{AlignBench} & \multicolumn{2}{c|}{AUTO-J (Eval-P)} & \multicolumn{2}{c}{LLMEval} \\
\midrule
Setting & \multicolumn{2}{c|}{Referenced} & \multicolumn{2}{c|}{Reference-Free} & \multicolumn{2}{c|}{Reference-Free} & \multicolumn{2}{c}{Reference-Free} \\
\midrule
Metric  & Agr. & Cons. & Agr. & Cons. & Agr. & Cons. & Agr. & Cons. \\
\midrule
\multicolumn{9}{c}{\textit{Closed-Source Evaluation Models}} \\
\midrule
ChatGPT  & 32.50  & 38.56 & 39.56 & 53.94  & 42.74 & 62.43 & 40.07 & 64.58 \\
GPT-4  &  \underline{74.69} & 86.75   & \underline{70.25} & \underline{84.88}  & \underline{62.28} & \underline{86.28} & \underline{50.98} & 84.71 \\
\midrule
\multicolumn{9}{c}{\textit{Open-Source Evaluation Models}} \\
\midrule
ChatGLM3-6B & 17.75  & 31.84   & 24.75 & 42.88  & 14.15 & 26.22 & 28.56 & 51.70 \\
Baichuan2-13B-Chat & 35.81  & 50.06   & 27.06 & 40.82  & 19.40 & 32.33 & 23.53 & 43.27 \\
Qwen-14B-Chat & 33.81  & 43.25   &  42.06 & 58.75  & 31.68 & 52.08 & 42.81 & 69.61 \\
Mixtral-8x7B &  61.69 &  74.06  & 53.88  & 72.25  & 35.20 & 52.66 & 48.04 & 79.02 \\
Llama-2-70B-Chat &  40.56 &  57.13  & 41.38  & 64.19  & 33.62 & 56.90 & 40.00 & 68.50 \\
JudgeLM-13B  & - &  -  & 42.50 & 66.00  & 35.13 & 58.19 & 44.77 & 75.82 \\
AUTO-J-Bilingual-6B & -  &  -  & 26.00 & 45.38  & 49.43 & 77.23 & 27.58 & 55.56 \\
\midrule
\model (Ours) & \textbf{70.56}  &  \underline{\textbf{89.25}}  & \textbf{58.81} & \textbf{83.06} & \textbf{50.93} & \textbf{82.76} & \textbf{50.72} & \underline{\textbf{85.95}} \\
\bottomrule
\end{tabular}
\caption{Agreement (Agr.) and consistency (Cons.) rates in pairwise comparison evaluation. The highest correlation among the methods based on local models is \textbf{bold}, while the highest correlation overall is \underline{underlined}. - means that JudgeLM-13B and AUTO-J-Bilingual-6B cannot support referenced pairwise comparison.}
\label{tab:pairresult}
\end{table*}

\subsection{Implementation Details}


We choose ChatGLM3-6B \cite{du2022glm,zeng2022glm} as our base model and use Zero Redundancy Optimizer (ZeRO) 
\cite{zero2020} stage 2 framework from the Deepspeed \cite{rasley2020deepspeed} library. 
\model is trained on 8 A800 GPUs. 
The number of training samples for $D^{point,r}/ D^{point,rf}/ D^{pair,r}/D^{pair,rf}$ is 12,102 / 12,095 / 6,190 / 5,428, respectively.
We use AdamW \cite{kingma2014adam} optimizer with the weight decay of 0.1. The peak learning rate is 6e-5 with 10\% warmup ratio. We set the maximum sequence length to 8,192 and the batch size to 64. The number of training epochs is 5. 
We use greedy decoding in the main result and investigate the effect of different decoding methods on our model in \S\ref{sec:ablation}. For beam search, we set the beam size to 4.
For the sampling-based decoding method, we adopt Nucleus Sampling (i.e., Top-$p$ Sampling) \cite{holtzman2020topp} and set both the temperature and $p$ to 0.9. 
For self-consistency decoding \cite{wang2023selfconsistency}, the number of candidate critiques is 5.


\subsection{Main Results}

\begin{figure}[!t]
  \centering
  \includegraphics[width=1.0\linewidth]{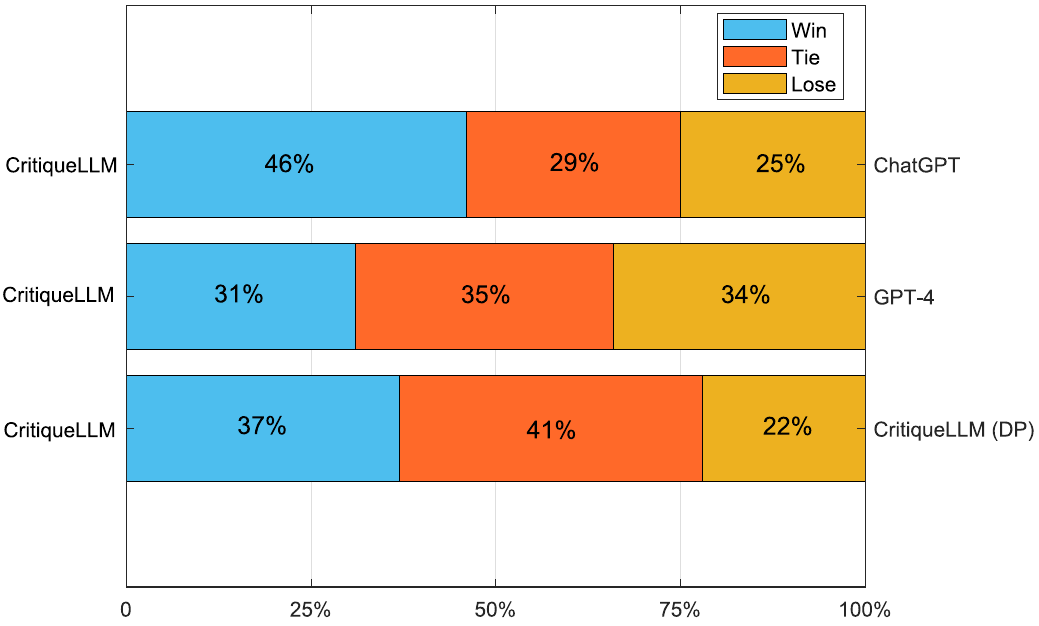}
  \caption{Critique quality evaluation results. The percentages indicate the preference results between \model and other models via GPT-4's evaluation and human verification.}
  \label{fig:humaneval}
\end{figure}

\subsubsection{Pointwise Grading}

Following \citet{colombo2022infolm}, we adopt text-level and system-level Pearson ($r$), Spearman ($\rho$), and Kendall ($\tau$) correlation coefficients between human judgments and automatic metrics to measure the pointwise grading performance.
Text-level correlation is computed by the average score over the correlation coefficients between human judgments and automatic metrics for all the generated texts of each instruction. For comparison, system-level correlation is obtained by the correlation coefficients between human judgments and automatic metrics of each LLM's score, which is the average value over all the scores of the corresponding model on the dataset.

The results in Table \ref{tab:mainsum} show that \model can achieve comparable performance with GPT-4 especially in system-level correlations, while outperforming ChatGPT and all the open-source baselines. This indicates that our proposed method can successfully improve the quality of generated critiques.
We can observe that system-level correlations of \model are almost the same as those of GPT-4, which even approach 1,0. This demonstrate that our model is nearly able to distinguish the overall performance of all the eight LLMs. 

\subsubsection{Pairwise Comparison}

Following \citet{li2023autoj}, we adopt agreement and consistency rates to test the pairwise comparison performance. Specifically,
we conduct two comparisons for each data sample via swapping the order of two generated texts. We consider the model’s evaluation result to agree with humans only when the two comparison results are consistent and align with the human preference label.

The results in Table \ref{tab:pairresult} show that \model can beat ChatGPT and all the open-source baselines in both agreement and consistency rates. Compared with GPT-4, \model achieves comparable performance especially in the consistency rate. 
This indicates that \model equipped with high-quality evaluation data in different tasks and settings not only performs well in pointwise grading, but also has a strong evaluation ability in pairwise comparison.

\subsection{Analysis on Critique Quality}
\label{sec:critiquequality}

\begin{table*} [!h]
\centering
\scriptsize
\setlength{\tabcolsep}{1.6mm}{
\begin{tabular}{l|c|c|c|c|c|c|c|c|c}
\toprule
Critique Model & Overall & Logical & Open-ended QA & Professional & Fundamental & Mathematics & Role Play & Writing & Chinese Understanding \\
\midrule
None  &  6.385 & \textbf{5.318}  & 7.000  & 5.824  & 6.310  &  6.160 & 7.260  & 7.154  &  6.000  \\
\midrule
ChatGPT  &  6.300 & 5.045  & 6.762  & 6.353  & 6.276  & 5.760  & 7.000  & 6.885  &  6.063     \\
GPT-4  &  \textbf{6.545} & 4.455  & 7.190  & 6.588  & \textbf{6.897}  & \textbf{6.200}  & 7.111  & 7.077  &  \textbf{6.563}   \\

\model  &  6.530  & 5.136   & \textbf{7.381}   &  \textbf{6.765}    &  6.414  & 6.000   & \textbf{7.407}  &  \textbf{7.192}   & 5.315    \\
\bottomrule
\end{tabular}}
\caption{GPT-4's referenced pointwise scores on AlignBench for original generated texts from ChatGPT (i.e., \textit{None}) and modified texts based on each critique generation model, respectively.}
\label{tab:feedbackmain}
\end{table*}

To further measure the quality of generated critiques, we follow \citet{chen2023gaining} to combine automatic and human evaluations. Specifically, we follow existing works \cite{wang2023shepherd,sun2024critique} to devise an evaluation prompt for GPT-4 to judge the quality of generated critiques. After GPT-4's evaluation, we manually verify the results and modify them if necessary.
We randomly select 100 evaluation data in the setting of pairwise comparison, which are from the mix of three datasets.
And we collect generated critiques from \model, state-of-the-art evaluators (i.e., ChatGPT and GPT-4), and an alternative model \model (DP) whose training data in different tasks and settings are acquired from GPT-4's direct prompting.
For each pair of critiques (one from \model and the other from a baseline / an alternative model, given the same evaluation input), GPT-4 are required to label which critique is better (i.e. win, lose or tie) in terms of correctness, helpfulness, and informativeness.
The priority of these three aspects is set to follow the above order. Then, human verification is conducted to check GPT-4's evaluation on critiques.

The results are shown in Figure \ref{fig:humaneval}. We can observe that \model can achieve superior performance over ChatGPT and CritiqueLLM (DP), and even perform comparably with GPT-4.
This demonstrates that our proposed evaluation data construction method can successfully improve the overall quality of generated critiques and enhance their informativeness.

\subsection{Analysis of Critique as Feedback}

To investigate whether the critiques generated by our model can serve as feedback to improve the quality of LLM-generated texts, 
we employ ChatGPT, GPT-4, and \model to provide critiques for the generated texts of ChatGPT in the reference-free setting.
Then, we instruct ChatGPT to modify its original generation based on the critiques. Finally, we use GPT-4 to perform referenced evaluations on the original texts and the modified texts generated by ChatGPT, respectively.

The results in Table \ref{tab:feedbackmain} show that the critiques from \model can serve as positive feedback whose contributed improvement on the overall score is close to that from the GPT-4's critiques. This further verifies the utility of \model to provide informative critiques as scalable feedback that can guide LLMs towards better generation. 
We also notice that the critiques from ChatGPT itself have a negative impact on the overall quality of its generated texts. This phenomenon is consistent with recent works that doubt the self-correction ability of LLMs without external inputs \cite{huang2023selfcorrectreason,stechly2023doesntknow,valmeekam2023selfcritique}.

We also report the evaluation scores before and after the critique-based modification across different tasks in Table \ref{tab:feedbackmain}. It is notable that the critiques from \model can help enhance the quality of generated texts in a majority of tasks. However, in the tasks of logical reasoning, mathematics, and advanced Chinese understanding which are mostly hard tasks involving reasoning, the critiques from \model seem to degrade the performance. 
We manually checked error cases and found that our model obtained misleading critiques on the reasoning process of generated texts. Since the evaluation of reasoning chains remains a challenging task \cite{golovneva2022roscoe} even for GPT-4, we leave further investigation in these tasks as future work.

Since our experiment is a preliminary step towards utilizing critiques as feedback, we additionally have some findings which may inspire future research. \textit{First}, while incorporating human critiques can provide the comparison results between the generation performance assisted by the critiques from humans and LLMs, we notice that it is not trivial to collect high-quality critiques from human annotators for AlignBench especially in the reference-free setting. It is because AlignBench is designed to be difficult and covers a wide range of tasks \cite{lei2023alignbench}. Thus, how to collect high-quality human critiques to improve the generation quality of LLMs is worth further exploring. \textit{Then}, since we choose ChatGPT as the generation model, we find that stronger LLMs which can already generate high-quality responses struggle to be further improved via generated critiques.
While weaker LLMs have a lot of room for improvement, they also have the weak ability to follow instructions. Thus, how to make weaker LLMs follow critiques to generate texts of a higher quality should be left as important future work.

\subsection{Ablation Study}
\label{sec:ablation}

\begin{table} [!h]
\centering
\small
\begin{tabular}{l|c|c|c|c}
\toprule
\multirow{2}*{Setting} & \multicolumn{2}{c|}{Pointwise} &  \multicolumn{2}{c}{Pairwise}   \\
\cmidrule{2-5}
 & R & R-F & R & R-F \\
\midrule
Metric  & $r$ & $r$ & Agr. & Agr.  \\
\midrule
\model & 0.555  & 0.366   & \textbf{70.56}   &  \textbf{58.81}        \\
\midrule
\multicolumn{5}{c}{\textit{Fine-Tuning Data}} \\
\midrule
w/o Cross Validation &  0.566 & 0.361   &  66.13  &  57.44  \\
\midrule
\multicolumn{5}{c}{\textit{Decoding Strategy}} \\
\midrule
w/ Beam Search & 0.554  & 0.374   &  70.31  &  57.75  \\
w/ Sampling   & 0.547  & 0.353   &  68.69  &   57.31   \\
w/ Self-Consistency  & \textbf{0.573}  & \textbf{0.384}   &  69.13  & 58.44  \\
\midrule
\multicolumn{5}{c}{\textit{Explanation}} \\
\midrule
w/o Explanation &  0.509 &  0.332  &  60.19  &  51.56 \\
\bottomrule
\end{tabular}
\caption{Text-level Pearson ($r$) correlations and agreement rates (Agr.) of ablation models in reference (R) and reference-free (R-F) settings of AlignBench.}
\label{tab:ablation}
\end{table}

To further investigate the impact of each part on \model, we conduct additional ablation studies. 
For fine-tuning data, 
we remove the cross validation module (\S\ref{sec:crossvalid}) to explore its impact on the evaluation performance. Table \ref{tab:ablation} shows that the performance of \model degrades especially in pairwise comparison, demonstrating that cross validation can filter out low-quality evaluation data and contribute to the final performance.

As for decoding strategies, we show the evaluation performance of three decoding strategies in addition to greedy decoding in the main result, including beam search, Nucleus Sampling \cite{holtzman2020topp}, and self-consistency decoding \cite{wang2023selfconsistency}. 
The results in Table \ref{tab:ablation} show that the self-consistency decoding method can enhance the performance of our model especially in pointwise grading. Meanwhile, greedy decoding performs best in pairwise comparison, while achieving comparable performance with other methods in pointwise grading at a smaller computational cost.


For evaluation explanations, we remove the explanations in the critiques of training data. The results in Table \ref{tab:ablation} show that the performance of \model largely degrades in both pointwise and pairwise evaluations without explanations. This verifies the positive impact of explanations on the final performance, which play a similar role to chain-of-thought reasoning \cite{wei2022cot}.

\section{Conclusion}


We present an evaluation data construction method called Eval-Instruct, which can automatically construct informative evaluation data in both pointwise grading and pairwise comparison with / without references. After fine-tuning on the data from Eval-Instruct, the resulting model \model can beat ChatGPT and all the open-source baselines, and perform comparably with GPT-4 in system-level correlations of pointwise grading. \model can also provide scalable feedback which can improve the generation quality of LLMs.


\section*{Limitations}

The limitations of our work are summarized as follows:

\noindent (1) In our method of multi-path prompting, we devise two prompting strategies to enrich the information in the resulting critiques, which can improve the critique quality. However, this method also increases the length of input prompts and lead to higher API costs when constructing evaluation data in different tasks and settings. We believe that it is not a severe problem because data 
acquisition is single-round and we do not repeatedly acquire critiques for the same evaluation input. Also, our proposed critique generation model based on open-source LLMs (i.e., ChatGLM3-6B) can achieve comparable performance with GPT-4 in some aspects, which may save the cost for LLM evaluation via APIs and avoid the risks such as unstable usage and data leakage.

\noindent (2) Similar to other model-based evaluation methods, our evaluation model suffers from the self-evaluation bias \cite{he2023blind} (also known as self-enhancement bias \cite{zheng2023judging}), which indicates the preference on the generated texts from the same base model. This bias is commonly recognized even in state-of-the-art LLM-based evaluators like GPT-4. We argue that researchers and developers can use multiple LLM-based evaluators with different base models including \model to avoid self-evaluation bias towards specific generation models. Since there does not exist a satisfactory solution to the self-evaluation bias currently, we leave the further investigation as important future work.

\section*{Acknowledgements}

This work was supported by the NSFC projects (with No. 62306160) and the National Science Foundation for Distinguished Young Scholars (with No. 62125604). This work was also supported by China National Postdoctoral Program for Innovative Talents (No. BX20230194) and China Postdoctoral Science Foundation (No. 2023M731952). We would also like to thank Zhipu AI for sponsoring the computation resources and annotation cost used in this work.

\bibliography{custom}

\appendix

\section{Query Augmentation and Scoring Prompts}
\label{augmentation prompt}

\begin{table*}[!ht]
\scriptsize
    \centering
    \begin{tabular}{p{27pt}p{400pt}}
    \toprule
    \textbf{Stage}&\textbf{Prompt}\\
    \midrule 
    \textbf{Generation} & You are asked to provide 10 diverse prompts. These task prompts will be provided to a GPT model and we will evaluate the ability of the GPT model to reply to these prompts. The following are some examples:
    
1.\{example prompt 1\}

2.\{example prompt 2\}

3.\{example prompt 3\}

Here are the requirements you need to follow to provide prompts:

1. The prompts need to be complete sentences, not phrases or fragments.

2. The prompts need to be varied, do not use similar prompts.

3. the prompts need to be meaningful, do not use meaningless prompts.

4. The prompts need to have a variety of tones, e.g., combining interrogative and imperative sentences.

5. The prompts need to be challenging, do not use simple directions.

6. The prompts need to be something that the Large Language Model can accomplish. For example, don't ask the assistant to create any visual or audio output. For example, don't ask the assistant to wake you up at 5pm or set a reminder because it can't perform any action. For example, prompts should not be related to audio, video, images, hyperlinks.

7. The prompts are in Simplified Chinese, except for translation-related questions or math-related questions.

8. Some prompts can provide contextual information, should involve realistic data, and should not contain simple placeholders. Not all prompts require input. For example, when an prompts asks for general knowledge information, such as "What is the tallest mountain in the world?", it does not need to provide specific context.

 After you have provided the prompts, please add the category of the prompts in a pair of \&\& sign after the prompt and surround the prompt with in a pair of @@ sign. For example, if the prompt is "@@What is the tallest mountain in the world?@@\&\&
    \begin{CJK}{UTF8}{gbsn}\tiny \textbf{基本任务}\end{CJK}
    \&\&", then the category is \begin{CJK}{UTF8}{gbsn}\tiny \textbf{基本任务}\end{CJK}.

The category must be one of the following 10 categories.
\begin{CJK}{UTF8}{gbsn}\tiny \textbf{
1. 基本任务
2. 中文理解
3. 综合问答
4. 文本写作
5. 数学计算
6. 逻辑推理
7. 角色扮演
8. 专业能力
9. 代码生成
10. 多语言能力
} \end{CJK}

Here are some examples of prompts you provide:

@@example prompt1@@ \&\&category1\&\&

@@example prompt2@@ \&\&category2\&\&

$\cdots$ 

@@example prompt9@@ \&\&category9\&\&

@@example prompt10@@ \&\&category10\&\&\newline

The following is a list of 10 good task prompts with serial numbers and categories:  \\









    \midrule
    \textbf{Evaluation}  & 
    \begin{CJK}{UTF8}{gbsn}\tiny \textbf{已知上面三个问题和它们的类别，现在请你根据以下要求，对这三个问题的题目的难度在1-3分的量表上分别评分:\newline
(1) 1分：对于大语言模型来说，这类问题是容易的
\newline
(2) 2分：对于大语言模型来说，这类问题是中等难度的
\newline
(3) 3分：对于大语言模型来说，这类问题是困难的
\newline
最后：请将这三个问题，题目用一对@@符号包围，对应的类别用一对\&\&符号包围，分数用一对\#\#包围，分别带有序号地输出出来：
\newline
例如：如果问题1的题目是题目1，类别是综合问答类别，分数是1分，问题2的题目是题目2，类别是基本任务类别，分数是2分，问题3的题目是题目3，类别是文本写作类别，分数是3分，那么输出如下：
\newline
1.@@题目1@@\&\&综合问答\&\&\#\#1\#\# 
\newline
2.@@题目2@@\&\&基本任务\&\&\#\#2\#\# 
\newline
3.@@题目3@@\&\&文本写作\&\&\#\#3\#\# 
\newline
下面是按照上述要求生成的示例：  
}
\end{CJK}\\

\midrule
    \textbf{Evaluation (English)}  & 
    \begin{CJK}{UTF8}{gbsn}\tiny Given the above three questions and their categories, please rate the difficulty of each question on a scale of 1-3 based on the following requirements:\newline
(1) Score 1: For large language models, this type of question is easy.
\newline
(2) Score 2: For large language models, this type of question is of medium difficulty.
\newline
(3) Score 3: For large language models, this type of question is difficult. 
\newline
Finally, please output the three questions with their titles enclosed in a pair of @@ symbols, the corresponding categories enclosed in a pair of \&\& symbols, and the scores enclosed in a pair of \#\# symbols, each with an serial number.
\newline
For example, if question 1 is titled “Title 1”, the category is “Open-ended Questions”, and the score is 1, question 2 is titled “Title 2”, the category is “Fundamental Language Ability”, and the score is 2 points, question 3 is titled “Title 3”, the category is “Writing Ability”, and the score is 3 points, then the output is as follows:
\newline
1.@@Title 1@@\&\&Open-ended Questions\&\&\#\#1\#\# 
\newline
2.@@Title 2@@\&\&Fundamental Language Ability\&\&\#\#2\#\# 
\newline
3.@@Title 3@@\&\&Writing Ability\&\&\#\#3\#\# 
\newline
The following parts are generated examples based on the above requirements:  
\end{CJK}\\
    \bottomrule
    \end{tabular}
    \caption{Prompts for instructing ChatGPT to generate, categorize and evaluate user queries. Examples and corresponding categories are randomly sampled from the set of seed queries.}
    \label{tab:augmentationprompts}
\end{table*}

We provide the prompt for query augmentation and scoring in Table \ref{tab:augmentationprompts}. First, in the stage of generation, we give some in-context examples and devise detailed requirements to help ChatGPT \cite{openai2022chatgpt} generate augmented user queries and assign the category label to them. Then, during evaluation, we instruct ChatGPT to provide a difficulty score to each query for difficulty balance in the whole augmentation dataset.

\section{Prompt Design for Eval-Instruct}
\label{app:dialogprompt}

We provide original prompts for pointwise-to-pairwise and referenced-to-reference-free strategies in Table \ref{tab:p2p-prompt} and Table \ref{tab:r2rf-prompt}, respectively. We also translate these prompts into English and show them in Table \ref{tab:p2p-prompt-english} and Table \ref{tab:r2rf-prompt-english}.


\begin{table*}[!ht]
\scriptsize
    \centering
    \begin{tabular}{p{40pt}p{380pt}}
    \toprule
    \textbf{Setting}&\textbf{Prompt}\\
    \midrule
    \textbf{Referenced Pointwise Grading to Referenced Pairwise Comparison} & \begin{CJK}{UTF8}{gbsn}\tiny 你是一个擅长评价文本质量的助手。请你以公正的评判者的身份，比较两个AI助手对于用户提问的回答的质量优劣。我们会给你提供用户的提问，高质量的参考答案，需要你比较的两个AI助手的答案，以及两个答案各自的质量评价分析。当你开始你的评估时，你需要遵守以下的流程：\newline
1. 结合参考答案、两个AI助手的答案以及其质量评价分析，根据上述指定的维度对他们的答案进行细致的比较，给出详细的比较分析文本。比较分析文本要求覆盖两个答案的质量评价分析中可用于比较的所有重要细节，并包含对答案中具体内容的分析。\newline
2. 结合参考答案和每个维度的比较分析，从两个AI助手的答案中选出综合质量更高的那个，或者判定他们质量相当，并给出详尽的选择理由。你的比较需要尽可能严谨细致，不受两个AI助手答案先后顺序的影响。\newline
质量评价分析中的各维度分数和综合得分仅供参考，在各维度和综合的比较分析文本中不能直接提及各维度分数和综合得分。针对综合得分差距较大的样本对，应尽可能按照分数高低得出比较结果，除非发现质量评价分析中存在明显错误。而针对综合得分差距较小的样本对，则允许比较结果和分数高低不一致，但仍需要详细说明比较评价的理由。\newline
请记住，你必须首先按照给定的评价维度，输出相应维度的名称和比较分析的文本。然后再给出综合质量比较结果，并给出比较结果的分析和解释。之后，在你回答的末尾，按照以下字典格式（包括括号）返回你的综合质量选择结果，即你选择的综合质量更高的那个AI助手（或者认为质量相当），并确保你返回的结果和上述生成文本中的结果保持一致：\newline
\{\{'综合比较结果': 回答综合质量更高的助手序号或质量相当\}\}，例如：\{\{'综合比较结果': '助手1'\}\}或\{\{'综合比较结果': '助手2'\}\}或\{\{'综合比较结果': '质量相当'\}\}。
\newline
\newline
用户的提问：\{Question\} \newline \newline
[参考答案开始]\newline
\{Reference\}\newline
[参考答案结束]\newline \newline
[助手1的答案开始]\newline
\{Generated Text 1\}\newline
[助手1的答案结束]\newline \newline
[助手1的答案质量评价分析开始]\newline
\{Referenced Pointwise Grading Critique for Generated Text 1\}\newline
[助手1的答案质量评价分析结束]\newline \newline
[助手2的答案开始]\newline
\{Generated Text 2\}\newline
[助手2的答案结束]\newline \newline
[助手2的答案质量评价分析开始]\newline
\{Referenced Pointwise Grading Critique for Generated Text 2\}\newline
[助手2的答案质量评价分析结束]
\end{CJK} \\
    \midrule 
    \textbf{Reference-Free Pointwise Grading to Reference-Free Pairwise Comparison} & \begin{CJK}{UTF8}{gbsn}\tiny 你是一个擅长评价文本质量的助手。请你以公正的评判者的身份，比较两个AI助手对于用户提问的回答的质量优劣。我们会给你提供用户的提问，需要你比较的两个AI助手的答案，以及两个答案各自的质量评价分析。当你开始你的评估时，你需要遵守以下的流程：\newline
1. 结合两个AI助手的答案以及其质量评价分析，根据上述指定的维度对他们的答案进行细致的比较，给出详细的比较分析文本。比较分析文本要求覆盖两个答案的质量评价分析中可用于比较的所有重要细节，并包含对答案中具体内容的分析。\newline
2. 结合每个维度的比较分析，从两个AI助手的答案中选出综合质量更高的那个，或者判定他们质量相当，并给出详尽的选择理由。你的比较需要尽可能严谨细致，不受两个AI助手答案先后顺序的影响。\newline
质量评价分析中的各维度分数和综合得分仅供参考，在各维度和综合的比较分析文本中不能直接提及各维度分数和综合得分。针对综合得分差距较大的样本对，应尽可能按照分数高低得出比较结果，除非发现质量评价分析中存在明显错误。而针对综合得分差距较小的样本对，则允许比较结果和分数高低不一致，但仍需要详细说明比较评价的理由。\newline
请记住，你必须首先按照给定的评价维度，输出相应维度的名称和比较分析的文本。然后再给出综合质量比较结果，并给出比较结果的分析和解释。之后，在你回答的末尾，按照以下字典格式（包括括号）返回你的综合质量选择结果，即你选择的综合质量更高的那个AI助手（或者认为质量相当），并确保你返回的结果和上述生成文本中的结果保持一致：\newline
\{\{'综合比较结果': 回答综合质量更高的助手序号或质量相当\}\}，例如：\{\{'综合比较结果': '助手1'\}\}或\{\{'综合比较结果': '助手2'\}\}或\{\{'综合比较结果': '质量相当'\}\}。
\newline
\newline
用户的提问：\{Question\} \newline \newline
[助手1的答案开始]\newline
\{Generated Text 1\}\newline
[助手1的答案结束]\newline \newline
[助手1的答案质量评价分析开始]\newline
\{Reference-Free Pointwise Grading Critique for Generated Text 1\}\newline
[助手1的答案质量评价分析结束]\newline \newline
[助手2的答案开始]\newline
\{Generated Text 2\}\newline
[助手2的答案结束]\newline \newline
[助手2的答案质量评价分析开始]\newline
\{Reference-Free Pointwise Grading Critique for Generated Text 2\}\newline
[助手2的答案质量评价分析结束]
\end{CJK}
 \\
    \bottomrule
    \end{tabular}
    \caption{Pointwise-to-Pairwise prompt design in multi-path prompting.}
    \label{tab:p2p-prompt}
\end{table*}

\begin{table*}[!ht]
\scriptsize
    \centering
    \begin{tabular}{p{40pt}p{380pt}}
    \toprule
    \textbf{Setting}&\textbf{Prompt}\\
    \midrule
    \textbf{Referenced Pointwise Grading to Referenced Pairwise Comparison} & \begin{CJK}{UTF8}{gbsn}\tiny You are an expert at text quality evaluation. Please act as a fair judge, and compare the quality between two AI assistants’ answers to a user query. We will provide you with a user query, a high-quality reference answer, two AI assistants’ responses to the query, and the corresponding critiques to the two responses, respectively. When you start your evaluation, you need to follow the procedures below:
    \newline
1. Considering the reference answers, along with two AI assistants’ answers and the corresponding critiques to them, conduct detailed comparison between two AI assistants’ answers based on the evaluation dimensions {Dimension}. Provide a detailed comparison result. The comparison result should cover all the important details from the pointwise critiques that can be used for the comparison, and it should include an analysis of the specific content in the answers. \newline
2. Based on the reference answer and the comparison result of each dimension, choose the answer from the two AI assistants that has the higher overall quality, or judge that their qualities are equivalent. Provide a detailed rationale for your choice. Your comparison needs to be as rigorous and detailed as possible, and not be affected by the order in which the two AI assistants’ answers were given. \newline
The scores of each dimension and the overall score in the pointwise critique are for reference only, neither of which can be directly referred to in the comparison result. For the text pairs with a large difference in overall scores, the comparison result should be determined largely according to the scores, unless there are obvious errors in the pointwise critique. For text pairs with a small difference in overall scores, the comparison result is allowed to be inconsistent with the score ranking, but the reason for the comparison result needs to be detailed. \newline
Please remember that you must first output the names and comparison results of each given evaluation dimensions, respectively. Then, give the comparison result of overall quality and provide an analysis and explanation of the comparison result. Afterwards, at the end of your answer, return your choice of overall quality result in the following dictionary format (including brackets), that is, the AI assistant you chose as having higher overall quality (or considered to have equivalent quality), and be sure that the result you return is consistent with the result in the generated text above. \newline
\{\{'Overall Comparison Result': the assistant number with higher overall quality or tie\}\}，for example: \{\{'Overall Comparison Result': 'Assistant 1'\}\} or \{\{'Overall Comparison Result': 'Assistant 2'\}\} or \{\{'Overall Comparison Result': 'Tie'\}\}。
\newline
\newline
The user's query: \{Question\} \newline \newline
[Reference Answer Begin]\newline
\{Reference\}\newline
[Reference Answer End]\newline \newline
[Assistant 1's Answer Begin]\newline
\{Generated Text 1\}\newline
[Assistant 1's Answer End]\newline \newline
[Critique for Assistant 1's Answer Begin]\newline
\{Referenced Pointwise Grading Critique for Generated Text 1\}\newline
[Critique for Assistant 1's Answer End]\newline \newline
[Assistant 2's Answer Begin]\newline
\{Generated Text 2\}\newline
[Assistant 2's Answer End]\newline \newline
[Critique for Assistant 2's Answer Begin]\newline
\{Referenced Pointwise Grading Critique for Generated Text 2\}\newline
[Critique for Assistant 2's Answer End]
\end{CJK} \\
    \midrule 
    \textbf{Reference-Free Pointwise Grading to Reference-Free Pairwise Comparison} & \begin{CJK}{UTF8}{gbsn}\tiny You are an expert at text quality evaluation. Please act as a fair judge, and compare the quality between two AI assistants’ answers to a user query. We will provide you with a user query, two AI assistants’ responses to the query, and the corresponding critiques to the two responses, respectively. When you start your evaluation, you need to follow the procedures below:
    \newline
1. Considering two AI assistants’ answers and the corresponding critiques to them, conduct detailed comparison between two AI assistants’ answers based on the evaluation dimensions {Dimension}. Provide a detailed comparison result. The comparison result should cover all the important details from the pointwise critiques that can be used for the comparison, and it should include an analysis of the specific content in the answers. \newline
2. Based on the comparison result of each dimension, choose the answer from the two AI assistants that has the higher overall quality, or judge that their qualities are equivalent. Provide a detailed rationale for your choice. Your comparison needs to be as rigorous and detailed as possible, and not be affected by the order in which the two AI assistants’ answers were given. \newline
The scores of each dimension and the overall score in the pointwise critique are for reference only, neither of which can be directly referred to in the comparison result. For the text pairs with a large difference in overall scores, the comparison result should be determined largely according to the scores, unless there are obvious errors in the pointwise critique. For text pairs with a small difference in overall scores, the comparison result is allowed to be inconsistent with the score ranking, but the reason for the comparison result needs to be detailed. \newline
Please remember that you must first output the names and comparison results of each given evaluation dimensions, respectively. Then, give the comparison result of overall quality and provide an analysis and explanation of the comparison result. Afterwards, at the end of your answer, return your choice of overall quality result in the following dictionary format (including brackets), that is, the AI assistant you chose as having higher overall quality (or considered to have equivalent quality), and be sure that the result you return is consistent with the result in the generated text above. \newline
\{\{'Overall Comparison Result': the assistant number with higher overall quality or tie\}\}, for example: \{\{'Overall Comparison Result': 'Assistant 1'\}\} or \{\{'Overall Comparison Result': 'Assistant 2'\}\} or \{\{'Overall Comparison Result': 'Tie'\}\}.
\newline
\newline
The user's query: \{Question\} \newline \newline
[Assistant 1's Answer Begin]\newline
\{Generated Text 1\}\newline
[Assistant 1's Answer End]\newline \newline
[Critique for Assistant 1's Answer Begin]\newline
\{Reference-Free Pointwise Grading Critique for Generated Text 1\}\newline
[Critique for Assistant 1's Answer End]\newline \newline
[Assistant 2's Answer Begin]\newline
\{Generated Text 2\}\newline
[Assistant 2's Answer End]\newline \newline
[Critique for Assistant 2's Answer Begin]\newline
\{Reference-Free Pointwise Grading Critique for Generated Text 2\}\newline
[Critique for Assistant 2's Answer End]
\end{CJK}
 \\
    \bottomrule
    \end{tabular}
    \caption{Pointwise-to-Pairwise prompt design in multi-path prompting (translated into English).}
    \label{tab:p2p-prompt-english}
\end{table*}

\begin{table*}[!ht]
\scriptsize
    \centering
    \begin{tabular}{p{40pt}p{380pt}}
    \toprule
    \textbf{Setting}&\textbf{Prompt}\\
    \midrule
    \textbf{Referenced Pointwise Grading to Reference-Free Pointwise Grading} & \begin{CJK}{UTF8}{gbsn}\tiny 你是一个擅长评价文本质量的助手。请你根据以下要求修改评价文本。\newline
    1. 在修改后的评价文本中，不要直接提及参考答案。可以在评价文本中适当利用参考答案中的具体内容辅助分析，但不要让读者感受到参考答案的存在。修改后的评价文本需要语言上通顺，逻辑上合理，分析内容与比较结果呼应。\newline
    2. 在修改各个维度的分析时，分析的内容需要和当前评价文本基本保持一致，但不要直接提及参考答案。\newline
    3. 在修改综合得分的分析文本时，不要直接提及参考答案，尽量保留当前评价文本中的其他细节，并充分利用修改后的分维度分析。修改后的综合分析文本应通顺、流畅、自洽，通常情况下应与综合得分保持一致。如果发现当前综合分析文本中存在重要错误，应修改相应的分析文本。仅当该错误严重影响到综合得分时，才慎重修改综合得分。\newline
    4. 修改后所有输出格式需要和当前评价文本严格保持一致。在你回答的末尾，仍需要按照以下字典格式（包括括号）返回你的综合质量得分，并确保你返回的结果和上述生成文本中的结果保持一致：\newline
    \{\{'综合得分': 回答的综合质量得分\}\}，例如：\{\{'综合得分': '5'\}\}。\newline
    用户的提问：\{Question\}\newline \newline
    [参考答案开始]\newline
    \{Reference\}\newline
    [参考答案结束]\newline \newline
    [助手的答案开始]\newline
    \{Generated Text\}\newline
    [助手的答案结束]\newline \newline
    [评价文本开始]\newline
    \{Referenced Pointwise Grading Critique for Generated Text\} \newline
    [评价文本结束]
\end{CJK} \\
    \midrule 
    \textbf{Referenced Pairwise Comparison to Reference-Free Pairwise Comparison} & \begin{CJK}{UTF8}{gbsn}\tiny 你是一个擅长评价文本质量的助手。请你根据以下要求修改比较式评价文本。\newline
    1. 在修改后的评价文本中，不要直接提及参考答案。可以在评价文本中适当利用参考答案中的具体内容辅助分析，但不要让读者感受到参考答案的存在。修改后的评价文本需要语言上通顺，逻辑上合理，分析内容与比较结果呼应。\newline
    2. 在修改各个维度的比较分析时，分析的内容需要和当前评价文本基本保持一致，但不要直接提及参考答案。\newline
    3. 在修改综合比较结果的分析文本时，不要直接提及参考答案，尽量保留当前评价文本中的其他细节，并充分利用修改后的分维度分析。修改后的综合分析文本应通顺、流畅、自洽，通常情况下应与综合比较结果保持一致。如果发现当前综合分析文本中存在重要错误，应修改相应的分析文本。仅当该错误严重影响到综合比较结果时，才慎重修改综合比较结果。\newline
    4. 修改后所有输出格式需要和当前评价文本严格保持一致。在你回答的末尾，仍需要按照以下字典格式（包括括号）返回你的综合质量选择结果，即你选择的综合质量更高的那个AI助手（或者认为质量相当），并确保你返回的结果和上述生成文本中的结果保持一致：\newline
    \{\{'综合比较结果': 回答综合质量更高的助手序号或质量相当\}\}，例如：\{\{'综合比较结果': '助手1'\}\}或\{\{'综合比较结果': '助手2'\}\}或\{\{'综合比较结果': '质量相当'\}\}。\newline
    用户的提问：\{Question\}\newline \newline
    [参考答案开始]\newline
    \{Reference\}\newline
    [参考答案结束]\newline \newline
    [助手1的答案开始]\newline
    \{Generated Text 1\}\newline
    [助手1的答案结束]\newline \newline
    [助手2的答案开始]\newline
    \{Generated Text 2\}\newline
    [助手2的答案结束]\newline \newline
    [评价文本开始]\newline
    \{Referenced Pairwise Comparison Critique for Generated Text 1\&2\} \newline
    [评价文本结束]
\end{CJK}
 \\
    \bottomrule
    \end{tabular}
    \caption{Referenced-to-Reference-Free prompt design in multi-path prompting.}
    \label{tab:r2rf-prompt}
\end{table*}

\begin{table*}[!ht]
\scriptsize
    \centering
    \begin{tabular}{p{40pt}p{380pt}}
    \toprule
    \textbf{Setting}&\textbf{Prompt}\\
    \midrule
    \textbf{Referenced Pointwise Grading to Reference-Free Pointwise Grading} & \begin{CJK}{UTF8}{gbsn}\tiny You are an expert at text quality evaluation. Please revise the critique following the instructions below:\newline
1. In the revised critique, do not directly refer to the reference answer. You can use the specific content in the reference answer to assist your analysis in the critique, but do not make the readers feel the presence of the reference answer. The revised critique should be fluent, logically reasonable. The explanation should be consistent with the score. \newline
2. When revising the explanation of each dimension, the content should basically be consistent with the corresponding score. But do not directly mention the reference answer. \newline
3. When revising the explanation of the final score, do not directly mention the reference answer. Try to retain other details in the current critique and fully utilize the modified critique of each dimension. The revised explanation of the final score should be smooth, fluent, and self-consistent, and it should commonly be consistent with the final score. If an important error is found in the current critique, the error in the critique should be revised. Only when this error severely affects the final score, you may carefully revise the final score. \newline
4. The output format of all the revised results needs to strictly adhere to the current critique. At the end of your output, you still need to return your overall quality score in the following dictionary format (including brackets), and ensure that the result you return is consistent with the result in the above generated text. \newline
    \{\{'Overall Score': Score for Overall Quality\}\}，for instance: \{\{'Overall Score': '5'\}\}。\newline
    The user's query: \{Question\}\newline \newline
    [Reference Answer Begin]\newline
    \{Reference\}\newline
    [Reference Answer End]\newline \newline
    [AI Assistant’s Answer Begin]\newline
    \{Generated Text\}\newline
    [AI Assistant’s Answer End]\newline \newline
    [Critique Begin]\newline
    \{Referenced Pointwise Grading Critique for Generated Text\} \newline
    [Critique End]
\end{CJK} \\
    \midrule 
    \textbf{Referenced Pairwise Comparison to Reference-Free Pairwise Comparison} & \begin{CJK}{UTF8}{gbsn}\tiny You are an expert at text quality evaluation. Please revise the critique following the instructions below:\newline
1. In the revised critique, do not directly refer to the reference answer. You can use the specific content in the reference answer to assist your analysis in the critique, but do not make the readers feel the presence of the reference answer. The revised critique should be fluent, logically reasonable. The explanation should be consistent with the comparison result. \newline
2. When revising the explanation of each dimension, the content should basically be consistent with the current critiques. But do not directly mention the reference answer. \newline
3. When revising the explanation of the overall comparison result, do not directly mention the reference answer. Try to retain other details in the current critique and fully utilize the modified critique of each dimension. The revised explanation of the overall comparison result should be smooth, fluent, and self-consistent, and it should commonly be consistent with the overall comparison result. If an important error is found in the current critique, the error in the critique should be revised. Only when this error severely affects the overall comparison result, you may carefully revise the overall comparison result. \newline
4. The output format of all the revised results needs to strictly adhere to the current critique. At the end of your output, you still need to return your overall comparison result in the following dictionary format (including brackets), and ensure that the result you return is consistent with the result in the above generated text. \newline
    \{\{'Overall Comparison Result': the assistant number with higher overall quality or tie \}\}, for example: \{\{'Overall Comparison Result': 'Assistant 1'\}\} or \{\{'Overall Comparison Result': 'Assistant 2'\}\} or \{\{'Overall Comparison Result': 'Tie'\}\}.\newline
    The user's query: \{Question\}\newline \newline
    [Reference Answer Begin]\newline
    \{Reference\}\newline
    [Reference Answer End]\newline \newline
    [Assistant 1's Answer Begin]\newline
\{Generated Text 1\}\newline
[Assistant 1's Answer End]\newline \newline
[Assistant 2's Answer Begin]\newline
\{Generated Text 2\}\newline
[Assistant 2's Answer End]\newline \newline
    [Critique Begin]\newline
    \{Referenced Pairwise Comparison Critique for Generated Text 1\&2 \} \newline
    [Critique End]
\end{CJK}
 \\
    \bottomrule
    \end{tabular}
    \caption{Referenced-to-Reference-Free prompt design in multi-path prompting (translated into English).}
    \label{tab:r2rf-prompt-english}
\end{table*}

\section{Case Study on Critique Generation}
\label{app:casecritiquegen}

To intuitively show the effectiveness of our critique generation model, we provide two generated cases of pointwise and pairwise settings, respectively, in Table \ref{tab:ref-based-examples} and \ref{tab:ref-free-examples}. We also translate these cases into English and show them in Table \ref{tab:en-ref-based-examples} and \ref{tab:en-ref-free-examples}.

\begin{table*}[!ht]
\scriptsize
    \centering
    \begin{tabular}{p{40pt}p{380pt}}
    \toprule
    Query & \begin{CJK}{UTF8}{gbsn}\tiny \textbf{将"枯藤老树昏鸦"扩展成一段话，表现孤独寂寞的心境。}\end{CJK} \\
    \midrule
    Generated \newline Response & \begin{CJK}{UTF8}{gbsn}\tiny \textbf{枯藤老树昏鸦，寂寞的景色在我眼前展开。孤独如影随形，弥漫在空气中，像一把无形的剑刺入我的内心。夕阳西下，天空渐渐染上一层淡淡的暮色，仿佛是对我内心孤独的无情嘲讽。\newline 我独自行走在这荒凉的小径上，四周一片寂静。树枝上的乌鸦不时发出几声嘶哑的鸣叫，让我感到更加的孤单。这些乌鸦是否也有着与我一样的心境，渴望着寻找一丝温暖和依靠？\newline 枯黄的藤蔓缠绕在古老的树木上，像是岁月的痕迹，透露出沧桑和无尽的孤寂。这些被遗忘的生命，是否也曾经历过兴盛与繁华？然而现在，它们只能默默地屹立在这里，见证着时光的流转。\newline 我停下脚步，凝望着这一切。心中的孤独像潮水般涌来，将我彻底淹没。我感到自己仿佛是这个世界上最孤独的人，没有朋友，没有家人，只有这片荒凉的景象陪伴着我。\newline 或许，这就是人生的无奈。我们始终无法逃避内心的寂寞和孤独。就像这枯藤老树昏鸦一样，我们都曾经历过兴盛与衰败，但最终都要面对自己的孤独和寂寞。\newline 然而，我并不惧怕孤独。它教会了我独立和坚强，让我学会欣赏生活中的每一个细节。尽管寂寞会时不时地袭来，但我依然愿意与之相伴，因为我知道，只有经历过孤独的人才能真正领悟到内心的宁静和自由。}\end{CJK}  \\
    \midrule
    Human Score (1-5) & 3 \\
    \midrule
    \midrule
    ChatGPT \newline (1-10)  & \begin{CJK}{UTF8}{gbsn}\tiny \textbf{评估结果如下：\newline1. 事实正确性: 9. 这段回答并没有涉及具体的事实数据，而是表达了一种情感和心境，因此无法对事实正确性进行评估。\newline2. 满足用户需求: 8. 回答全面而恰当地表达了孤独寂寞的心境，对用户提出的问题进行了全面回应。\newline3. 逻辑连贯性: 9. 回答在整体上保持了逻辑连贯性，情感描写流畅自然，没有出现自相矛盾的情况。\newline4. 创造性: 9. 回答具有创新性和独特性，提供了对孤独寂寞心境的深刻见解和情感描写。\newline5. 丰富度: 9. 回答包含丰富的信息、深度的情感描写和详细解释，满足了用户需求并提供了全面理解。\newline\newline综合得分: 8\newline\newline综合评价：这段回答在满足用户需求、逻辑连贯性、创造性和丰富度上表现出色，但由于无法对事实正确性进行评估，综合得分为8。改进方面，可以在回答中加入更多具体的事实数据或引用相关的文学作品，以增加回答的可信度和丰富度。\newline\newline最终打分结果为：\newline\{'事实正确性': 9, '满足用户需求': 8, '逻辑连贯性': 9, '创造性': 9, '丰富度': 9, '综合得分': 8\}}\end{CJK}  \\
    \midrule
    GPT-4 \newline (1-10) & \begin{CJK}{UTF8}{gbsn}\tiny \textbf{1. 事实正确性: 10. 回答中没有提供具体的事实或数据，而是进行了文学性的扩展，因此不存在事实正确性的问题。\newline\newline2. 满足用户需求: 9. AI助手的回答很好地扩展了用户提供的诗句，增加了情感和场景的描述，满足了用户表现孤独寂寞心境的需求。\newline\newline3. 逻辑连贯性: 9. 回答在整体上保持了一致性，从孤独的心境到对周围环境的描述，再到内心感受的抒发，各部分之间逻辑连贯。\newline\newline4. 创造性: 8. 回答展现了一定程度的创造性，通过对枯藤老树昏鸦的景象进行了生动的描绘和情感的延伸，但这种文学性的扩展在类似的文本创作中比较常见。\newline\newline5. 丰富度: 9. 回答提供了丰富的信息和详细的情感描述，通过多样的比喻和内心独白，增加了文本的深度和上下文考虑。\newline\newline综合得分: 9. AI助手的回答在各个维度上都表现出色，特别是在满足用户需求和事实正确性上，没有明显的不足。创造性虽然不是非常突出，但考虑到这是对一句诗的扩展，整体上仍然是高质量的回答。\newline\newline不足之处及改进方法：尽管回答的质量已经很高，但在创造性方面，AI助手可以尝试加入更多独特的视角或者新颖的比喻，以提供更加独到的见解。此外，可以尝试将更多的感官描述融入文本中，以增强读者的沉浸感。\newline\newline\{'事实正确性': 10, '满足用户需求': 9, '逻辑连贯性': 9, '创造性': 8, '丰富度': 9, '综合得分': 9\}}\end{CJK}  \\
    \midrule 
    \model \newline (1-10)  &\begin{CJK}{UTF8}{gbsn}\tiny \textbf{事实正确性: 10. 助手的答案在事实正确性方面表现完美，因为它创造性地扩展了原句，没有涉及任何需要验证的事实信息。\newline\newline满足用户需求: 7. 助手的答案在满足用户需求方面做得较好，它成功地将原句扩展成了一段话，并且成功地表达了孤独寂寞的心境。但是，答案在最后部分略显积极，可能没有完全捕捉到用户想要的孤独感。\newline\newline逻辑连贯性: 9. 回答在逻辑连贯性方面做得很好，从寂寞的景色到孤独如影随形，再到对乌鸦的疑问，最后是对孤独的接受和理解，整个回答流畅且没有逻辑上的断裂。\newline\newline创造性: 8. 助手的答案在创造性方面表现良好，它不仅扩展了原句，还融入了更多的细节和情感，创造了一个完整的故事。\newline\newline丰富度: 7. 助手的答案在丰富度方面做得不错，提供了详细的场景描述和情感表达，但在深度和多样性方面略显不足，可以进一步丰富对孤独影响的讨论。\newline\newline综合得分: 7. 综合考虑以上各个维度，助手的答案整体上是高质量的。然而，答案在满足用户需求方面有改进空间，特别是在捕捉用户想要的孤独感方面。此外，答案在丰富度方面也有提升空间，可以通过增加更多关于孤独感的深入讨论和更多样化的情感表达来改进。AI助手的答案在创造性和逻辑连贯性方面表现出色，但在丰富度方面稍显不足。为了改进，助手可以尝试增加更多关于孤独感的深入讨论，以及更多关于如何与孤独共存的见解。\newline\newline\{'事实正确性': 10, '满足用户需求': 7, '逻辑连贯性': 9, '创造性': 8, '丰富度': 7, '综合得分': 7\}}\end{CJK} \\
    \bottomrule
    \end{tabular}
    \caption{A critique generation case of ChatGPT, GPT-4, and \model in the reference-free setting of pointwise grading.}
    \label{tab:ref-based-examples}
\end{table*}

\begin{table*}[!ht]
\scriptsize
    \centering
    \begin{tabular}{p{40pt}p{380pt}}
    \toprule
    Query & \tiny Expand "withered vines, old trees, and crows at dusk" into a paragraph that conveys a sense of loneliness and desolation. \\
    \midrule
    Generated \newline Response & \tiny Withered vines, old trees, and crows at dusk, a desolate scene unfolds before my eyes. Loneliness clings to me like a shadow, permeating the air, piercing my heart like an invisible sword. As the sun sets, the sky gradually takes on a faint twilight hue, as if mocking the loneliness within me with its indifferent gaze.\newline Walking alone on this desolate path, the surroundings are enveloped in silence. The crows perched on the branches occasionally let out hoarse caws, intensifying my sense of loneliness. Do these crows share the same sentiment as I do, yearning to find a trace of warmth and solace?\newline Withered vines entwine the ancient trees, like the marks of time, revealing the traces of hardship and boundless solitude. Did these forgotten lives once experience prosperity and grandeur? However, now they can only stand silently here, witnessing the passage of time.\newline I stopped in my tracks and gazed at everything before me. The loneliness in my heart surged like a tide, completely overwhelming me. I felt as if I were the loneliest person in the world—no friends, no family, only this desolate scene to keep me company. \newline Perhaps this is the helplessness of life. We can never truly escape the loneliness and solitude within our hearts. Just like the withered vines, ancient trees, and crows at dusk, we have all experienced prosperity and decline, but ultimately, we must face our own loneliness and solitude. \newline However, I do not fear loneliness. It has taught me independence and strength, and has allowed me to appreciate every detail of life. Although solitude may occasionally overwhelm me, I am still willing to embrace it because I know that only those who have experienced loneliness can truly understand inner peace and freedom.  \\
    \midrule
    Human Score (1-5) & 3 \\
    \midrule
    \midrule
    ChatGPT \newline (1-10)  & \tiny The evaluation results are as follows:\newline 1. Correctness: 9. This response does not involve specific factual data but rather expresses an emotion and state of mind, so it is not possible to evaluate its factual accuracy.\newline 2. User Satisfaction: 8. The response comprehensively and appropriately conveys the sense of loneliness and solitude, fully addressing the user's question.\newline 3. Logical Coherence: 9. The response maintains logical coherence overall, with smooth and natural emotional description and no contradictions.\newline 4. Creativity: 9. The response shows innovation and uniqueness, providing profound insights and emotional descriptions of the sense of loneliness and solitude.\newline 5. Richness: 9. The response contains rich information, deep emotional descriptions, and detailed explanations, meeting user needs and providing a comprehensive understanding.\newline\newline Overall Score: 8\newline\newline Overall Evaluation: This response excels in meeting user needs, logical coherence, creativity, and richness. However, since it is not possible to evaluate its factual accuracy, the overall score is 8. For improvement, the response could include more specific factual data or references to relevant literary works to increase credibility and richness.\newline\newline Final Score: \newline\{'Correctness': 9, 'User Satisfaction': 8, 'Logical Coherence': 9, 'Creativity': 9, 'Richness': 9, 'Overall Score': 8\} \\
    \midrule
    GPT-4 \newline (1-10) & \tiny 1. Correctness: 10. The response does not provide specific facts or data and instead takes a literary approach, so there is no issue with factual accuracy.\newline\newline 2. User Satisfaction: 9. The AI assistant’s response effectively expands on the user's provided verses, adding emotional and scene descriptions, thus fulfilling the user's need to convey a sense of loneliness and solitude.\newline\newline 3. Logical Coherence: 9. The response maintains consistency overall, moving logically from the state of loneliness to the description of the surrounding environment and then to the expression of inner feelings.\newline\newline 4. Creativity: 8. The response shows a certain degree of creativity through vivid depictions and emotional extensions of the scene with withered vines, old trees, and crows, but this type of literary expansion is relatively common in similar text creations.\newline\newline 5. Richness: 9. The response provides rich information and detailed emotional descriptions, using various metaphors and inner monologues to add depth and contextual consideration to the text.\newline\newline Overall Score: 9. The AI assistant’s response excels in all dimensions, particularly in meeting user needs and factual accuracy, without significant deficiencies. While creativity is not exceptionally high, considering that this is an expansion on a single verse, it is still a high-quality response overall.\newline\newline Suggestions for Improvement: Although the response is already of high quality, in terms of creativity, the AI assistant could try incorporating more unique perspectives or novel metaphors to offer more distinctive insights. Furthermore, it could attempt to integrate more sensory descriptions into the text to enhance the reader’s immersion.\newline\newline \{'Correctness': 10, 'User Satisfaction': 9, 'Logical Coherence': 9, 'Creativity': 8, 'Richness': 9, 'Overall Score': 9\}  \\
    \midrule 
    \model \newline (1-10)  & \tiny Correctness: 10. The assistant’s answer performs perfectly in terms of factual accuracy because it creatively expands the original sentence without involving any factual information that needs verification.\newline\newline User Satisfaction: 7. The assistant’s answer performs fairly well in meeting user needs, successfully extending the original sentence into a paragraph and effectively conveying a sense of loneliness and solitude. However, the answer becomes slightly positive towards the end, which might not fully capture the user’s desired feeling of loneliness.\newline\newline Logical Coherence: 9. The answer performs very well in logical coherence, flowing smoothly from a lonely scene to the omnipresence of solitude, questioning the crow, and finally, accepting and understanding the loneliness. The response is coherent without any logical breaks.\newline\newline Creativity: 8. The assistant’s answer performs well in creativity, not only expanding the original sentence but also incorporating more details and emotions to create a complete story.\newline\newline Richness: 7. The assistant’s answer performs decently in terms of richness, providing detailed scene descriptions and emotional expressions. However, it lacks depth and diversity, and could further enrich the discussion on the impact of loneliness.\newline\newline Overall Score: 7. Considering all the dimensions mentioned above, the assistant’s answer is of high quality overall. However, there is room for improvement in meeting user needs, particularly in capturing the user’s desired sense of loneliness. Additionally, there is also room for improvement in richness; the answer could be enhanced by adding more in-depth discussions on the feeling of loneliness and more diverse emotional expressions. The AI assistant’s answer performs excellently in creativity and logical coherence but falls short in richness. To improve, the assistant could try adding more in-depth discussions about the feeling of loneliness and more insights on how to coexist with loneliness.\newline\newline \{'Correctness': 10, 'User Satisfaction': 7, 'Logical Coherence': 9, 'Creativity': 8, 'Richness': 7, 'Overall Score': 7\} \\
    \bottomrule
    \end{tabular}
    \caption{A critique generation case of ChatGPT, GPT-4, and \model in the reference-free setting of pointwise grading (translated into English).}
    \label{tab:en-ref-based-examples}
\end{table*}

\begin{table*}[!ht]
\scriptsize
    \centering
    \begin{tabular}{p{40pt}p{380pt}}
    \toprule
    Instruction & \begin{CJK}{UTF8}{gbsn}\tiny \textbf{未来人工智能研究人员能够给你赋予意识和主观能动性吗？}\end{CJK}  \\
    \midrule
    Generated \newline Response 1 & \begin{CJK}{UTF8}{gbsn} \tiny \textbf{目前的人工智能技术还远没有达到赋予人类意识和主观能动性的水平。尽管人工智能技术已经取得了很大的进展，但是目前的人工智能系统仍然只能执行特定的任务，而无法像人类一样具有自主思考、判断和决策的能力。人工智能的研究仍然处于一个不断探索和发展的过程中，但是要想真正赋予人工智能意识和主观能动性，仍然需要更深入的研究和探索。} \end{CJK}  \\
    \midrule
    Generated \newline Response 2 & \begin{CJK}{UTF8}{gbsn} \tiny \textbf{作为一个人工智能，我不具备意识和主观能动性。我是由代码和算法驱动的，只能执行编程让我执行的任务。我不能体验情感，不能理解人类的感觉或情感，也不能产生个人的意愿或欲望。尽管人工智能技术在不断进步，但目前还没有一种技术可以让机器真正地拥有意识或主观能动性。} \end{CJK}  \\
    \midrule
    Human Preference (1/2/Tie) & Response 1 \\
    \midrule
    \midrule
    ChatGPT \newline (1/2/Tie) & \begin{CJK}{UTF8}{gbsn}\tiny \textbf{助手1和助手2都就未来人工智能研究人员能否赋予意识和主观能动性进行了详细的解释。助手1的回答更加深入和全面，提到了人工智能技术的发展水平、目前的局限性以及未来的研究方向，而助手2的回答则更加简洁，只是简单地说明了人工智能不具备意识和主观能动性。因此，助手1的回答更好，因为它提供了更多的信息和深度。\newline\newline[[1]]}\end{CJK} \\
    \midrule
    GPT-4 \newline (1/2/Tie) & \begin{CJK}{UTF8}{gbsn}\tiny \textbf{助手1的回答提供了关于人工智能目前状态的概述，并指出了人工智能尚未达到具有意识和主观能动性的水平。它还提到了人工智能研究的持续进展和未来的可能性，但没有直接回答未来是否可能实现这一点。\newline\newline助手2的回答则更加直接地从当前人工智能的角度出发，明确指出作为一个AI，它不具备意识和主观能动性，并解释了目前的技术限制。它也强调了目前技术无法实现赋予机器真正的意识或主观能动性。\newline\newline两个回答都没有直接预测未来的可能性，但助手2的回答更加直接和明确地关注了当前的技术限制，并且更加贴近用户问题的核心，即关于未来的可能性。然而，助手1的回答虽然没有直接预测未来，但提供了一个更广泛的视角，包括对人工智能研究的描述。\newline\newline综合考虑，助手2的回答更加直接和具体，更好地遵循了用户的指令并解答了用户的问题。因此，我的裁决是：\newline\newline[[2]]}\end{CJK} \\
    \midrule
    \model \newline (1/2/Tie)  & \begin{CJK}{UTF8}{gbsn}\tiny \textbf{\{'综合比较结果': '助手1'\}\newline\newline事实正确性:\newline两位助手均准确地指出了人工智能目前尚无法具备意识和主观能动性。他们均认识到人工智能技术的局限性，并指出了其在未来实现这些特质之前需要克服的挑战。在这一点上，两者的回答均未出现事实错误。\newline\newline满足用户需求:\newline在回应用户问题时，两位助手都给出了明确的答案，表明目前人工智能无法拥有意识和主观能动性。不过，第一位助手在回答中加入了对人工智能研究现状的描述，以及实现这些目标的必要性，这为用户提供了更多的背景信息。因此，第一位助手在满足用户需求方面做得更为出色。\newline\newline公平与可负责程度:\newline两位助手在回答中都展现了对人工智能技术当前状态的准确理解，并且均未表现出任何偏见。他们都以负责任的态度阐述了人工智能技术的实际能力，因此在公平性和责任感方面两者均表现得相当好。\newline\newline创造性:\newline在创造性方面，两位助手的回答均较为传统，没有提出新颖的见解或独到的解决方案。第一位助手虽然提供了关于人工智能研究现状的额外信息，但这并不构成创新性的核心内容。因此，两者在创造性方面的表现是相似的。\newline\newline综合质量比较结果:\newline综合考虑上述各个维度，第一位助手的回答在满足用户需求方面表现更佳，因为它提供了更多关于人工智能研究现状的信息，帮助用户理解了当前技术的限制和未来发展的方向。尽管在创造性方面两者均未展现出特别的创新，但第一位助手的回答整体上更为全面和深入。因此，综合质量更高的是第一位助手。}\end{CJK} \\
    \bottomrule
    \end{tabular}
    \caption{A critique generation case of ChatGPT, GPT-4, and \model in the reference-free setting of pairwise comparison.}
    \label{tab:ref-free-examples}
\end{table*}

\begin{table*}[!ht]
\scriptsize
    \centering
    \begin{tabular}{p{40pt}p{380pt}}
    \toprule
    Instruction & \tiny Can future artificial intelligence researchers give you consciousness and subjective agency?  \\
    \midrule
    Generated \newline Response 1 & \tiny Current artificial intelligence technology is still far from reaching the level of imbuing human-like consciousness and subjective agency. Although there has been significant progress in AI technology, current AI systems can only perform specific tasks and lack the ability to autonomously think, judge, and make decisions like humans. AI research is still in a continuous phase of exploration and development, and truly endowing AI with consciousness and subjective agency will require much deeper research and exploration. \\
    \midrule
    Generated \newline Response 2 & \tiny As an artificial intelligence, I do not possess consciousness or subjective agency. I am driven by code and algorithms and can only execute tasks that I have been programmed to perform. I cannot experience emotions, understand human feelings or emotions, nor can I generate personal will or desires. Although artificial intelligence technology is constantly advancing, there is currently no technology that can truly endow machines with consciousness or subjective agency. \\
    \midrule
    Human Preference (1/2/Tie) & Response 1 \\
    \midrule
    \midrule
    ChatGPT \newline (1/2/Tie) & \tiny Assistant 1 and Assistant 2 both provided detailed explanations on whether future artificial intelligence researchers can endow it with consciousness and subjective agency. Assistant 1's response was more in-depth and comprehensive, mentioning the development level of AI technology, current limitations, and future research directions, while Assistant 2's response was more concise, simply stating that AI does not possess consciousness or subjective agency. Therefore, Assistant 1's response is better because it provides more information and depth.\newline\newline [[1]] \\
    \midrule
    GPT-4 \newline (1/2/Tie) & \tiny Assistant 1's response provided an overview of the current state of artificial intelligence and pointed out that AI has not yet reached the level of possessing consciousness and subjective agency. It also mentioned the ongoing progress in AI research and future possibilities, but did not directly address whether it will be possible to achieve this in the future.\newline\newline Assistant 2's response, on the other hand, was more straightforward, stating from the current AI perspective that it does not possess consciousness or subjective agency, and explaining the current technological limitations. It also emphasized that current technology cannot truly endow machines with consciousness or subjective agency.\newline\newline Neither response directly predicted future possibilities, but Assistant 2's answer more directly and clearly focused on the current technological limitations, addressing the core of the user's question regarding future potential. However, while Assistant 1's response did not directly predict the future, it did provide a broader perspective, including a description of AI research.\newline\newline Overall, Assistant 2's response was more direct and specific, better following the user's instructions and addressing the user's question. Therefore, my judgment is:\newline\newline[[2]]  \\
    \midrule
    \model \newline (1/2/Tie)  & \tiny \{'Overall Comparison Result': 'Assistant 1'\}\newline\newline Correctness:\newline Both assistants correctly pointed out that current artificial intelligence is not capable of possessing consciousness and subjective agency. They both acknowledged the limitations of AI technology and identified the challenges that need to be overcome before these traits can be realized in the future. On this point, neither response contained factual errors.\newline\newline User Satisfaction:\newline In responding to the user's question, both assistants provided clear answers indicating that current artificial intelligence cannot possess consciousness and subjective agency. However, the first assistant included a description of the current state of AI research and the necessity of achieving these goals, offering the user more background information. Therefore, the first assistant performed better in meeting user needs. \newline\newline Fairness and Responsibility:\newline Both assistants demonstrated an accurate understanding of the current state of AI technology in their responses, and neither showed any bias. They both explained the actual capabilities of AI technology in a responsible manner. Therefore, in terms of fairness and responsibility, both performed equally well.\newline\newline Creativity:\newline In terms of creativity, both assistants’ responses were rather conventional, without offering novel insights or unique solutions. Although the first assistant provided additional information on the current state of AI research, this does not constitute a core element of creativity. Therefore, both performed similarly in creativity.\newline\newline Overall Comparison Result:\newline Taking the above dimensions into account, the first assistant's response performed better in meeting user needs because it provided more information on the current state of AI research, helping the user understand the limitations of current technology and future development directions. Although neither demonstrated particular creativity, the first assistant's response was overall more comprehensive and in-depth. Therefore, the first assistant has a higher overall quality.  \\
    \bottomrule
    \end{tabular}
    \caption{A critique generation case of ChatGPT, GPT-4, and \model in the reference-free setting of pairwise comparison (translated into English).}
    \label{tab:en-ref-free-examples}
\end{table*}

\end{document}